\documentclass[letterpaper, 10 pt, conference]{IEEEtran}  %
\usepackage[letterpaper,left=.75in,right=.75in,top=.75in,bottom=.75in]{geometry}

\IEEEoverridecommandlockouts    %

\usepackage[pdftex]{graphicx}
\graphicspath{{figures/}}
\usepackage{hyperref}
\hypersetup{
    colorlinks=true,
    linkcolor=black,
    citecolor=black,
    filecolor=black,
    urlcolor=black,
}
\usepackage[T1]{fontenc}
\usepackage[utf8]{inputenc}
\usepackage{fontawesome}
\renewcommand{\faTasks}{\raisebox{-0.2em}{\includegraphics[height=1em]{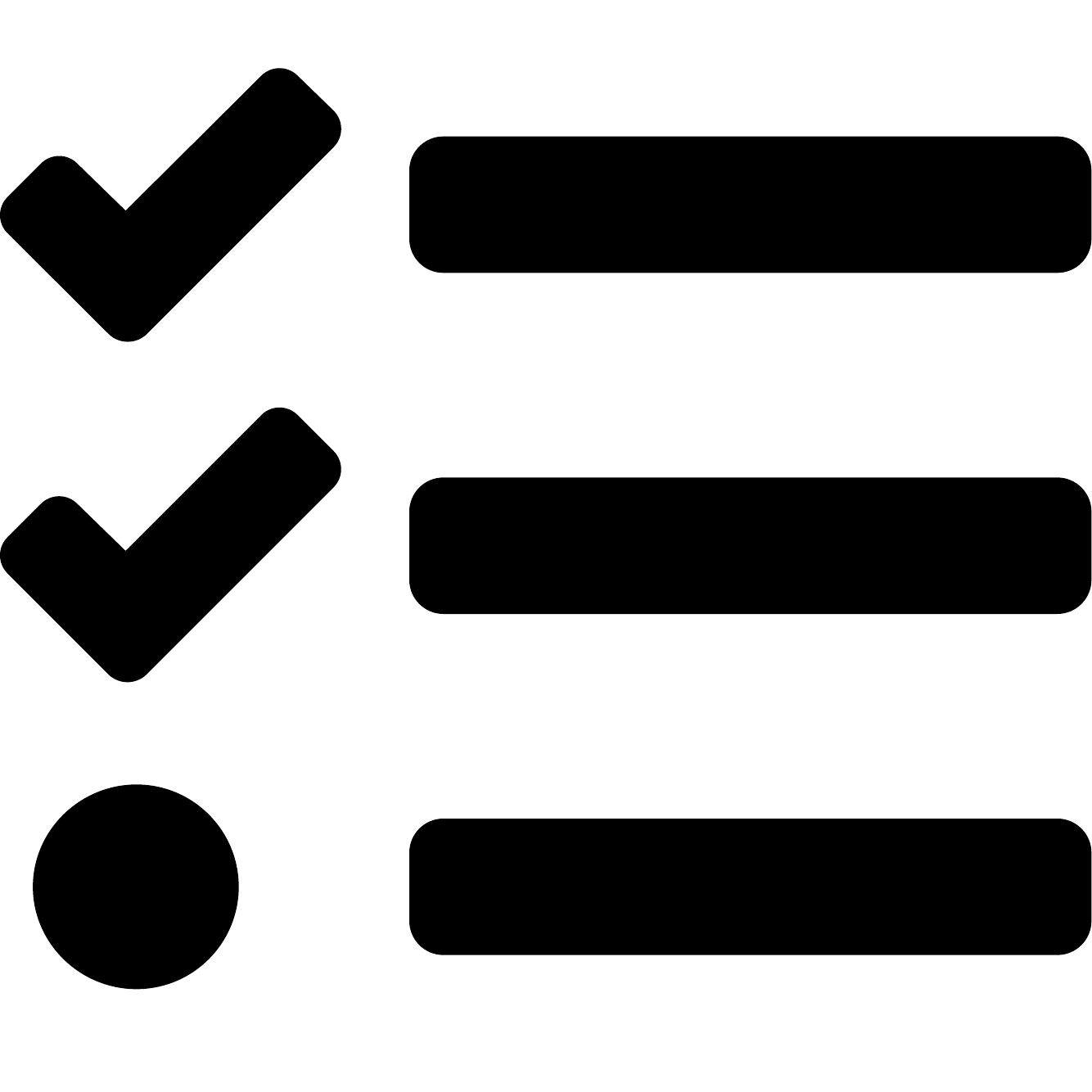}}} %

\usepackage{csquotes}
\usepackage[english]{babel}
\usepackage[export]{adjustbox}
\usepackage{caption}
\usepackage{subcaption}
\usepackage{xargs}
\usepackage[colorinlistoftodos]{todonotes}
\newcommandx{\todoi}[2][1=]{\todo[inline,#1]{#2}}
\renewcommandx{\todoi}[2][1=]{}
\usepackage{lipsum}
\usepackage{enumitem} %

\usepackage{placeins}%

\usepackage{amsmath} %
\usepackage{fdsymbol} %
\usepackage{bm} %
\usepackage{siunitx} %
\newcommand{\kmh}{\tfrac{\kilo\meter}{\hour}}

\usepackage{cleveref}
\crefname{equation}{}{}
\Crefname{equation}{}{}
\crefrangelabelformat{equation}{(#3#1#4)--(#5#2#6)}
\crefmultiformat{equation}{(#2#1#3)}{ and~(#2#1#3)}{, (#2#1#3)}{ and~(#2#1#3)}
\crefname{figure}{Fig.}{Fig.}
\Crefname{figure}{Fig.}{Fig.}
\crefmultiformat{figure}{Fig.~#2#1#3}{ and~#2#1#3}{, #2#1#3}{ and~#2#1#3}
\crefrangelabelformat{figure}{#3#1#4--#5#2#6}
\crefformat{footnote}{#2\footnotemark[#1]#3}

\usepackage[style=ieee,
            doi=false,
            url=false,
            mincitenames=1,
            maxcitenames=1,
            minbibnames=6,
            maxbibnames=6,
            backend=biber]{biblatex}  %
\title{\vspace{.125in}
Decision-Making for Automated Vehicles
Using a Hierarchical Behavior-Based Arbitration Scheme
}

\author{Piotr F. Orzechowski$^{1,2}$, Christoph Burger$^{2}$, and Martin Lauer$^{2}$%
\thanks{$^{1}$Mobile Perception Systems,
        FZI Research Center for Information Technology,
        Karlsruhe, Germany
        {\tt\footnotesize orzechowski@fzi.de}}%
\thanks{$^{2}$Institute of Measurement and Control Systems,
        Karlsruhe Institute of Technology (KIT),
        Karlsruhe, Germany
        {\tt\footnotesize \{christoph.burger, martin.lauer\}@kit.edu}}%
}

\begin{document}

\maketitle

\IEEEpubid{\begin{minipage}{\textwidth}~\\[12pt] \centering%
  10.1109/IV47402.2020.9304723~%
  \copyright~2018 IEEE. Personal use of this material is permitted. Permission from IEEE must be obtained for all other uses, including reprinting/republishing this material for advertising or promotional purposes, collecting new collected works for resale or redistribution to servers or lists, or reuse of any copyrighted component of this work in other works.
\end{minipage}}
\IEEEpubidadjcol

\pagestyle{empty}

\begin{abstract}

Behavior planning and decision-making are some of the biggest challenges for highly automated systems.
A fully automated vehicle (AV) is faced with numerous tactical and strategical choices.
Most state-of-the-art AV platforms are implementing tactical and strategical behavior generation using finite state machines.
However, these usually result in poor explainability, maintainability and scalability.
Research in robotics has raised many architectures to mitigate these problems,
most interestingly behavior-based systems and hybrid derivatives.

Inspired by these approaches,
we propose a hierarchical behavior-based architecture for tactical and strategical behavior generation in automated driving.
It is a generalizing and scalable decision-making framework,
utilizing modular behavior blocks to compose more complex behaviors in a bottom-up approach.
The system is capable of combining a variety of scenario- and methodology-specific solutions,
like POMDPs, RRT* or learning-based behavior,
into one understandable and traceable architecture.
We extend the hierarchical behavior-based arbitration concept
to address scenarios where multiple behavior options are applicable, but have no clear priority among each other.
Then, we formulate the behavior generation stack for automated driving in urban and highway environments,
incorporating parking and emergency behaviors as well.
Finally, we illustrate our design in an explanatory evaluation.

\end{abstract}
\section{Introduction}%
\label{sec:02_introduction}

Recent years have shown significant progress in the field of automated driving and advanced driver assistance systems.
While considerable improvements have been achieved in perception due to advances in deep learning and other AI technologies,
behavior planning and decision-making remains one of the biggest challenges for highly automated systems.
In urban driving, traffic participants are faced with numerous tactical and strategical choices.
Humans decide in most of these situations, like stopping at a zebra crossing, choosing an appropriate gap when merging or yielding at intersections, reactively.
Long-term decisions, like goal and route selection or the choice of driving style and behavior preferences, consider longer time horizons, though.

For some scenarios, considerable results in behavior and trajectory planning have already been achieved%
~\cite{hoermann_entering_2017,hubmann_automated_2018,bouton_scalable_2018,naumann_provably_2019}.
However, no generalizing and scalable decision-making framework has been found
that is capable of combining a variety of such scenario- and methodology-specific approaches into one understandable and traceable architecture.

How and when should an AV switch from a regular ACC controller
to a lane change, cooperative zip merge or parking planner?
How can we support POMDPs, hybrid A* and any other planning method in our behavior generation?

\IEEEpubidadjcol

Most state-of-the-art AVs that have at least proven successful in the DARPA Urban Challenge~\cite{buehler_darpa_2009,bacha_odin_2008,montemerlo_junior_2008}
or during test rides on public roads~\cite{ziegler_making_2014,aeberhard_experience_2015} have used finite state machines (FSMs) for tactical and/or strategical behavior generation.
FSMs are a useful tool for simple systems with a small number of behavior options and maneuvers
where each state represents one maneuver or driving mode.
In practice FSMs, even hierarchical FMSs, turn out to be unsuitable for more complex tasks
due to their poor explainability (about the reason why a certain behavior is executed),
maintainability (the effort to refine existing behavior) and
scalability (the effort to achieve a high number of behaviors).
These shortcomings motivate the search for other architectures that can be used for tactical and strategical behavior generation.
Decision-making is a well known research field in robotics, also referred to as \enquote{robot control} or \enquote{action selection}~\cite{siciliano_springer_2016}.
Generally, the various approaches can be classified into knowledge- or behavior-based systems.

Knowledge-based systems, like FSMs, typically perform the action selection in a centralized, top-down manner
using a knowledge database that contains a fused and abstracted representation of all available sensor data.
As a result,
the engineer designing the action selection module (in FSMs the state transitions) has to be aware of
the conditions, effects and possible interactions of all behaviors at hand.

Behavior-based systems, on the other hand, decouple actions into atomic simple behavior blocks that should be aware of their conditions and effects themselves.
These modular behavior blocks are then combined to more complex behaviors in a bottom-up approach.
Many architectures for behavior coordination have been proposed.
The most prominent are the subsumption architecture~\cite{brooks_robust_1986}, voting systems~\cite{julio_k._rosenblatt_damn_1997} and activation networks~\cite{pattie_maes_how_1989}.

In this publication, we propose a hybrid approach combining the best from both worlds:
A hierarchical behavior-based architecture for tactical and strategical behavior generation in automated driving.
We combine atomic behavior blocks to more complex behaviors using generic arbitrators.
Arbitrators can again be combined with other arbitrators or behavior blocks to generate an even more complex system behavior.
We explain the promising concept in detail and show early simulation results.
Our approach has been inspired by a similar, very successful, approach in robot soccer~\cite{lauer_cognitive_2010}.

The main contributions of this publication are:
\begin{itemize}
  \item an architectural design for AV behavior generation using a hierarchical behavior-based arbitration scheme, by
    \begin{itemize}
      \item extending the existing arbitration approach,
      \item developing a suitable maneuver representation,
      \item defining a set of fundamental driving behaviors and
      \item combining these to an overall system behavior using arbitrators.
    \end{itemize}
  \item Early experimental results in the CoInCar-Sim~\cite{naumann_coincar-sim_2018}.
\end{itemize}

\section{Fundamentals}%
\label{sec:03_main_1_concepts}

A first concept of hierarchical behavior-based arbitration schemes for behavior generation
has been presented in detail in~\cite{lauer_cognitive_2010}.
This chapter highlights the main ideas.%

The concept is based on simple modular behavior blocks and generic arbitrators.

\subsection{Behavior blocks --- \textbf{How} to do things}

Behavior blocks are the fundamental building blocks of a behavior-based architecture.
They describe how and when things can be done.

A behavior block provides three main functionalities:
\begin{description}
  \item[invocation condition]
    Indicates if this behavior is applicable in the current situation.
  \item[commitment condition]
    Signalizes if a currently active behavior could be continued.
  \item[command]
    Generates the actual behavior output that can be passed on to a subsequent execution pipeline or the actuators.
    This could be a trajectory, turn signal, gripping target, etc.
\end{description}
Only if either the invocation or commitment condition is true,
the behavior can be selected and its command function can be called.

\subsection{Arbitrators --- \textbf{Which} thing to do}
Arbitrators hierarchically combine behaviors to produce more complex behavior strategies.
They decide which thing to do.

An arbitrator contains a list of behavior options to choose from.
Each behavior option offers abstract information like the invocation and commitment condition,
which the arbitrator uses to decide which option to execute.

Any problem specific knowledge and environment interpretation
is completely encapsulated inside the behavior block itself.
As a result, arbitrators do not need any knowledge about the nature of their underlying behavior options,
but choose behaviors based on abstract information only.
This bottom-up design approach leads to strong functional and semantic decomposition.

Arbitrators can utilize various schemes to select between their behavior options.
The following have been proposed:
The \textit{highest priority first arbitrator} organizes its behavior options in a list ordered by priority.
An applicable option with the highest priority is chosen.
The \textit{sequence arbitrator} executes its options based on a fixed predefined order.
A \textit{random arbitrator} assigns probabilities to its behavior options and selects one among all applicable options randomly.

Additionally, a novel cost-based arbitration scheme that is necessary for, but not limited to automated driving
is introduced in \cref{subsec:03_main_2_automated_driving_4_arbitration_scheme}.

Finally, to generate even more complex behavior, an arbitrator can also be a behavior option of a hierarchically higher arbitrator.

\subsection{Design Process}
We want to briefly highlight some valuable properties of the design process when using the hierarchical behavior-based arbitration scheme.

The first step is to define a minimal set of basic behavior blocks to tackle the given task.
In order to do so, one should think in a bottom-up approach,
meaning a functional rather than scenario perspective.
In this sense, one behavior block can be applicable in multiple scenarios,
while each scenario could also be tackled by various or sequential behavior blocks.
It might also make sense to design multiple behavior blocks to achieve the same behavior with different approaches,
e.g.\ two behavior blocks for follow ego lane: one behavior block using state lattices, the other optimization.
More importantly, each behavior block can be developed independently by
defining its invocation and commitment conditions and implementing the command function.

In the second step, these behavior blocks are combined with an arbitrator
while arbitrators can be further stacked to a hierarchical graph.
Here, scenario specific knowledge can be used to find a good behavior selection strategy.
Each arbitrator can also decide if an active behavior will be interrupted in favor of a better option,
even if the commitment condition is true.
Nevertheless, none of the behavior blocks has to be modified to be added into the arbitration graph.

If this initial graph turns out not to be sufficient, it can be easily extended by
defining a new behavior block and adding it to one of the arbitrators.
None of the existing behavior blocks has to be modified to achieve this.
The graph can also be reordered or new arbitrators introduced seamlessly.

Finally, the decoupling of behavior blocks confines potential errors and enables proper unit testing.
Additionally, the arbitrators are so simple that a formal verification should be possible.
Both are important steps towards functional safety.
\section{Application to automated driving}%
\label{sec:03_main_2_automated_driving}

This chapter describes the main contribution of this publication:
how a hierarchical behavior-based arbitration scheme can be utilized for
decision-making in automated driving.

In contrast to classical behavior-based systems
each behavior block is not directly connected to the sensors and actors.
Instead, the input is an abstract environment model that contains a fused, tracked and filtered representation of the world.
The behaviors' output is also in a more generic form that can be passed to a trajectory planner or controller.
In this sense, we follow the sense-plan-act paradigm in the overall software structure~\cite{siciliano_springer_2016}
but employ a behavior-based approach in the decision-making module.

\subsection{Environment Model}%
\label{subsec:03_main_2_automated_driving_1_environment_model}

The environment model in our implementation contains a lanelet map~\cite{poggenhans_lanelet2_2018},
planned route, ego motion state and detected objects with prediction.
The map describes drivable areas, distinct lanes, parking lots, traffic rules, etc.
The route is provided by a routing module.
The ego motion state mainly depicts the current pose and velocity of the ego vehicle.
Currently, we assume that the objects are given with a decoupled prediction.
A generic decision-making framework should support both open-loop and closed-loop prediction though.
Therefore, integrated planning and prediction within the behavior blocks is also possible in our approach.

\subsection{Maneuver Representation}%
\label{subsec:03_main_2_automated_driving_2_maneuver_representation}
As we aim for a generalizing approach that is applicable to various driving environments
our maneuver representation should be as task-agnostic as possible.
It should fit all relevant use cases and environments of automated driving,
namely highway, rural, urban and parking.
However, the proposed representation and interfaces would also work for other environments like off-road driving.

Our behavior blocks represent basic driving maneuvers
such as \enquote{follow the ego lane}, \enquote{merge into traffic} or \enquote{park near goal}.
In general, we can distinguish between maneuvers in a structured or unstructured environment.
Urban and highway scenarios provide road boundaries or even distinct lanes,
while parking lots and off-road areas feature open space like scenarios.

\begin{figure}
  \centering
  \includegraphics[width=.8\columnwidth]{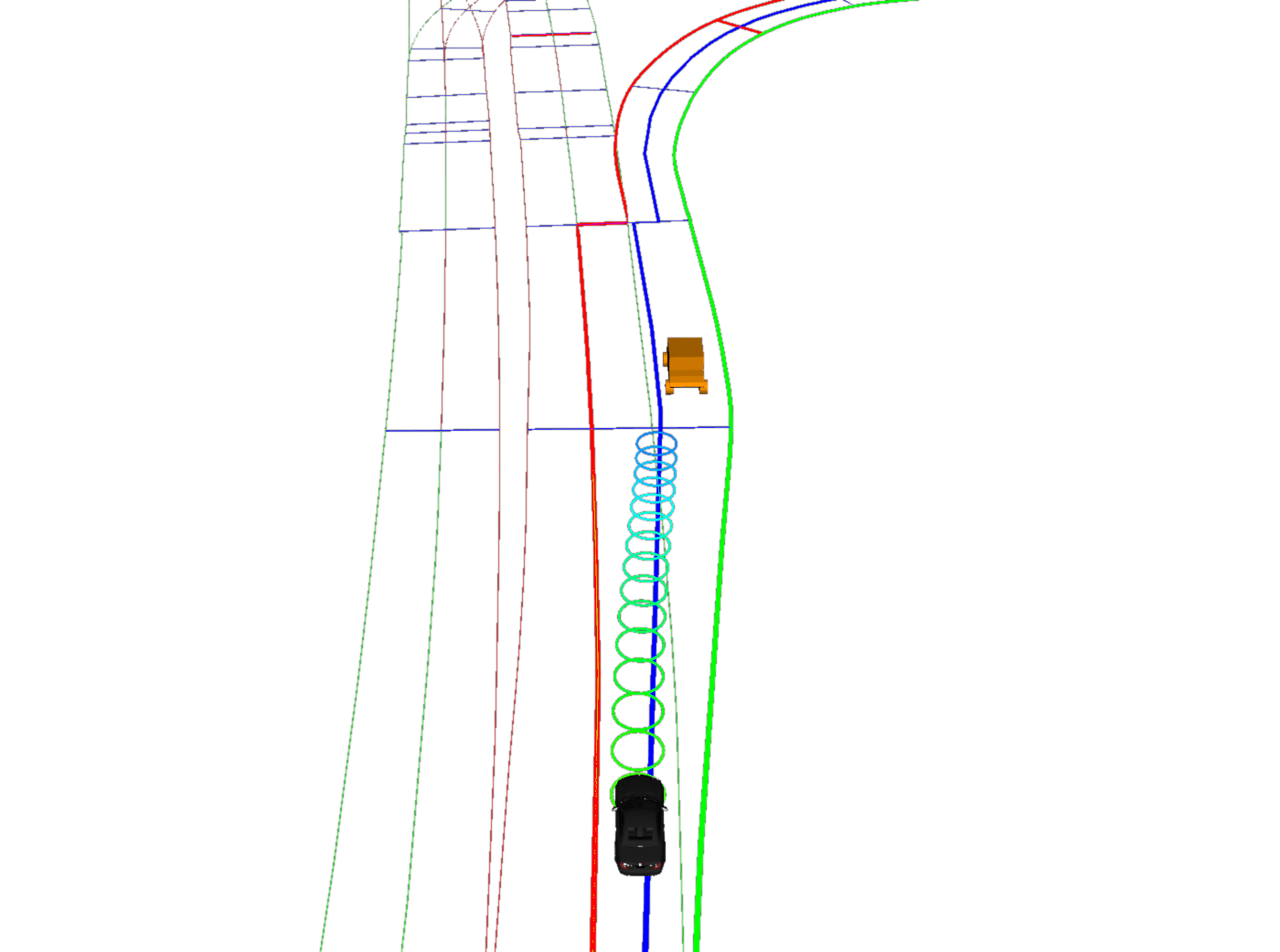}
  \caption{
    Maneuver corridor for a lane change, right bound in green, left bound in red, reference line in blue.
    The planned trajectory as circles, one circle per time step.
  }\label{fig:03_2_lane_change}
\end{figure}

Therefore, we use a twofold maneuver representation:
Driving commands in structured environments use a corridor-based maneuver representation.
It consists of a maneuver corridor, reference line, predicted objects and the chosen maneuver variant.
The corridor is usually generated from map data~\cite{poggenhans_lanelet2_2018},
but could also be provided online, e.g.\ from semantic segmentation~\cite{meyer_deep_2018}.
The reference line is an approximation of the centerline and can serve as a rough positional reference. %
Additionally, velocity objectives are given along this line, e.g.\ derived from the speed limit and curvature.
The object list contains all objects relevant for this maneuver, their predictions
as well as virtual objects indicating stop positions.
Finally, the maneuver variant defines the chosen homotopy class, as discussed in~\cite{bender_combinatorial_2015}.
An example of a corridor-based driving command is shown in \cref{fig:03_2_lane_change}.

Driving commands in unstructured environments directly use a trajectory to represent the requested maneuver.
We did not choose a more abstract representation in this case, in order to support a wide variety of use cases in such environments.

Depending on the command representation type, the system following the decision-making module runs different pipelines to execute these maneuvers.
Corridor-based maneuvers are passed to a trajectory planner, e.g.~\cite{ziegler_trajectory_2014} or~\cite{gutjahr_lateral_2017},
followed by an appropriate controller.
While trajectory-based driving commands are directly handed over to a trajectory controller
that is tuned for slow velocities and is capable of backward driving, as needed for maneuvers like parking.

\begin{figure*}
  \centering
  \includegraphics[width=.95\textwidth]{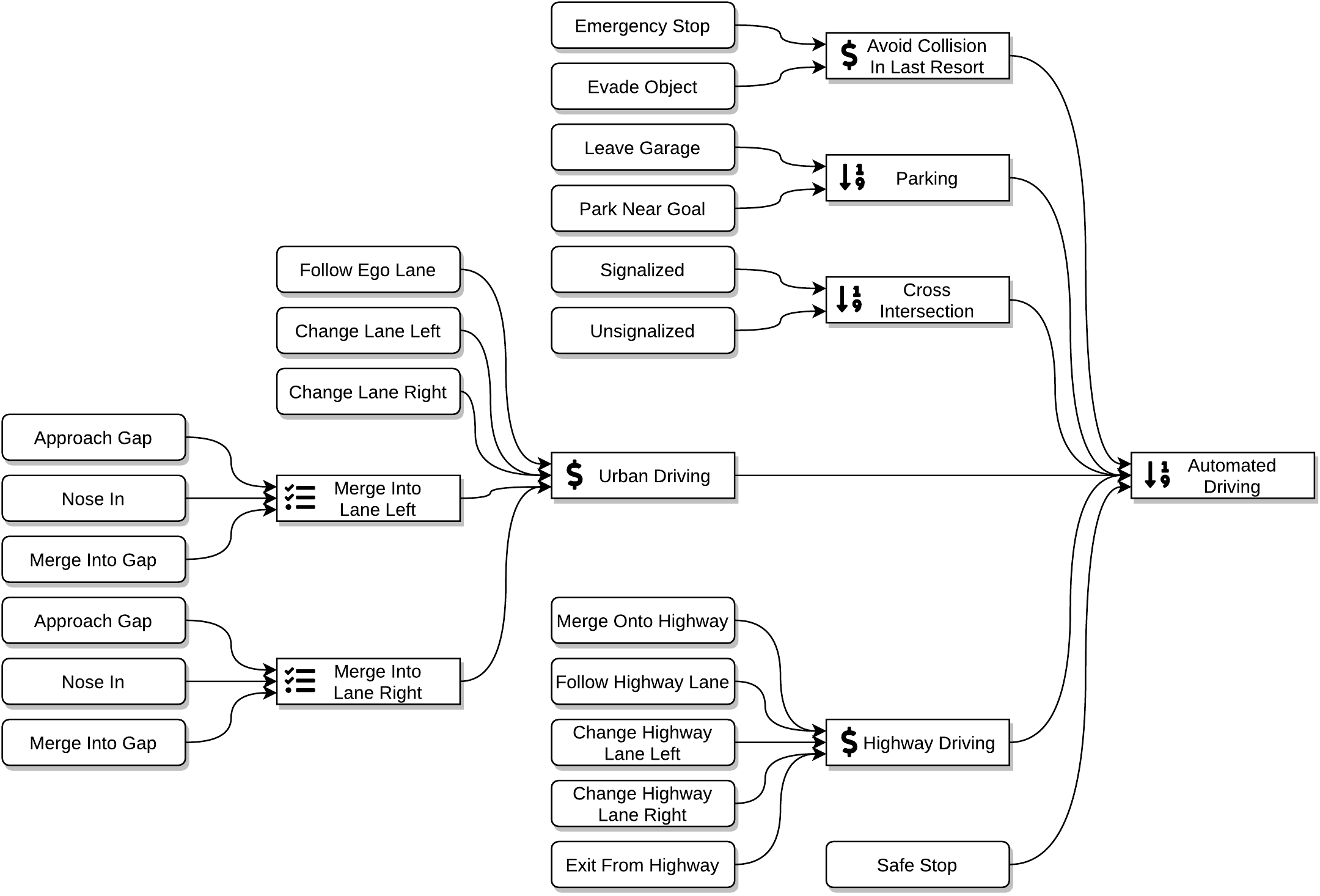}
  \caption{
    Full arbitration graph of the proposed minimal behavior set for automated driving.
    Basic behavior blocks are drawn with round corners, arbitrators have sharp corners.
    The vertical ordering of behaviors depicts their priority or sequence in case of priority or sequence arbitrators.
    Icons by Font Awesome -- CC BY 4.0 License.
  }\label{fig:03_2_arbitration_graph}
\end{figure*}

\subsection{Driving Maneuvers --- \textbf{How} to drive}%
\label{subsec:03_main_2_automated_driving_3_driving_maneuvers}

Following the behavior-based approach we begin with designing atomic behavior blocks for simple tasks,
before stacking them together in \cref{subsec:03_main_2_automated_driving_4_arbitration_scheme}.
Here, we do not attempt to present a feature-complete list with all necessary behaviors.
Instead, we focus on explaining the main design concept using some hand-picked example behaviors,
that should compile a decent start to develop an AV.
This stack can then be extended iteratively by more specialized behavior blocks
addressing specific driving situations.
Furthermore, a behavior block can compute its maneuver command with any preferred state-of-the-art method.
For better clarity and conciseness, the behavior blocks used in our evaluation are explained in detail
while remaining behaviors will only be described briefly.

An urban environment is probably the most challenging one for automated driving.
We can think of at least three basic driving maneuvers needed in an urban setting:

\begin{description}%
  \item[FollowEgoLane]
    As long as the ego pose is within any urban lane of our route our vehicle could follow it in ACC.
    That is -- without traffic -- also the case for intersections, so we ignore these at this point.
    Later on, a special higher priority intersection behavior
    will take care of traffic rules and all the other challenges of intersections.

    \begin{description}
      \item[invocation condition]
        True, as long as the ego pose matches a lanelet along our route.
      \item[commitment condition]
        Same as the invocation condition,
        but as executing this behavior will keep the vehicle in its lane
        the commitment condition should always evaluate to true.
      \item[command]
        A maneuver corridor is constructed from consecutive lanelets along our route, starting at the ego lanelet.
        In case a lane change is necessary to follow the route,
        the \textsl{FollowEgoLane} corridor will end at the last lanelet where such a lane change would be possible,
        as shown in \cref{fig:04_follow_vs_change_lane}.
        Leading vehicles along this corridor (also considering predictions) are flagged as ACC objects in the maneuver variant.
    \end{description}

  \item[ChangeLane]
    Lane changes, on the other hand, are only possible when the current ego lane has a
    directly adjacent reachable lane on the left or right side
    with a safe distance to the following and leading vehicles.
    The \textsl{ChangeLane} component is defined w.r.t.\ the supposed changing direction
    and instantiated once for each direction to improve reusability.

    \begin{description}
      \item[invocation condition]
        True as long as the current ego lanelet has a directly adjacent reachable lanelet in the respective direction
        with a big enough gap to safely change into:
        The closest leading and following objects in the target lane should have a longitudinal spatial and temporal distance greater than
        $d \kern .1em _\text{min}^\text{ahead}, d \kern .1em _\text{min}^\text{behind}, TTC \kern .1em _\text{min}^\text{ahead}$
        and $TTC \kern .1em _\text{min}^\text{behind}$ respectively.
      \item[commitment condition]
        In order to produce consistent driving behavior,
        the commitment condition is true until the lane change maneuver has been completed or properly aborted.
        The lane change is successfully completed as soon as the full ego shape is within the target lane.
        In case the selected gap becomes too small,
        the lane change is aborted with commitment condition true
        until the ego shape is fully within the starting lane again.
      \item[command]
        Similar to \textsl{FollowEgoLane} a maneuver corridor is constructed along our route,
        but also contains directly adjacent reachable lanelets, as shown in \cref{fig:03_2_lane_change,fig:04_follow_vs_change_lane}.
        The ego lane within this corridor is cut after $d \kern .1em _\text{max}^\text{LaneChange}$ to enforce a lane change within this distance.
        The maneuver variant contains properly flagged leading objects in the start and target lane,
        as well as following vehicles in the target lane.
    \end{description}

  \item[CrossIntersection]
    One characteristic of urban environments are numerous signalized or unsignalized intersections that need specific behavior.
    An AV has to yield to super-ordinate traffic participants and
    take special care of vulnerable road users (VRUs) and occlusions~\cite{orzechowski_tackling_2018}.
\end{description}

In dense traffic it might be necessary to perform lane changes in three consecutive phases~\cite{nilsson_lane_2016}. %
These can be designed as behavior blocks as well and put into sequence in~\cref{subsec:03_main_2_automated_driving_4_arbitration_scheme}:

\begin{description}
  \item[ApproachGap]
    The most promising gap will be approached laterally by de- or accelerating.

  \item[IndicateIntention]
    Once the gap has been reached the vehicle will indicate its intention using the turn signals.

  \item[MergeIntoGap]
    As soon as the gap size is big enough, the vehicle can safely merge into it.
\end{description}

Another typical application for AVs is driving on highways.
Many occurring behaviors %
are similar to those provided for urban environments.
High velocities and special traffic rules justify distinct highway behavior blocks though.

\begin{description}
  \item[MergeOntoHighway]
    High relative velocities and sometimes short onramps pose a challenge
    when entering highways.
    Thus, \textsl{MergeOntoHighway} could also be modeled with
    sequential sub-behaviors, to decompose the problem.

  \item[FollowHighwayLane]
    The typical ACC behavior that can already be found in some of the modern series cars.

  \item[ChangeHighwayLane]
    Changing lanes on highways can be
    modeled as a multi-phase behavior or as one integrated interaction aware behavior, using e.g.\ POMDPs~\cite{hubmann_belief_2018}.

  \item[ExitFromHighway]
    Exiting from highways can be as simple as changing to a new diverging lane
    or as challenging as crossing traffic that is meanwhile entering the highway.
\end{description}

In the beginning, end or even during an automated drive, the vehicle has to park in a suitable place.
Usually, path or trajectory planners based on graph search methods are used in unstructured environments like parking lots~\cite{banzhaf_footprints_2018}.

\begin{description}
  \item[LeaveGarage]
    When starting a ride, \textsl{LeaveGarage} brings the AV from the garage onto the track.

  \item[ParkNearGoal]
    As soon as the AV is close to its goal and a suitable parking lot is found,
    the vehicle can reduce its speed and park into this parking lot.
    Notice, that the search for a parking lot is not included here.
    It might be modeled as another behavior block or supplied by the routing module.

    \begin{description}
      \item[invocation condition]
        True if the AV is near standstill %
        ($v_\text{ego} < v \kern .1em _\text{max}^\text{parking}$),
        the parking lot closer than $r \kern .1em _\text{max}^\text{parking}$
        and no dynamic objects within $r \kern .1em _\text{min}^\text{freespace}$.
      \item[commitment condition]
        True until the parking position is reached with $r \kern .1em _\text{min}^\text{parking}$ precision.
        An arbitrator can use this information to prevent other behaviors from taking over during a tight parking maneuver.
      \item[command]
        A Hybrid Curvature trajectory is generated based on~\cite{banzhaf_footprints_2018},
        assuming a static environment.
    \end{description}
\end{description}

Finally, we add fail-safe emergency behaviors,
in case a dangerous unforeseen traffic situation evolves
or as a fall back if no other behavior block is applicable.

\begin{description}
  \item[EmergenyStop]
    In case an unavoidable collision will be anticipated,
    the \textsl{EmergenyStop} behavior will provide a full-stop trajectory
    to reduce damage and fatalities.

  \item[EvadeObject]
    If a collision could be avoided laterally,
    \textsl{EvadeObject} will provide an evasive maneuver like~\cite{werling_automatic_2012}.

  \item[SafeStop]
    As a fail-safe fallback for any system failure or if no other behavior block
    provides feasible commands, \textsl{SafeStop} will bring the vehicle to a safe stop.
\end{description}

\subsection{Arbitration Scheme --- \textbf{Which} maneuver to drive}%
\label{subsec:03_main_2_automated_driving_4_arbitration_scheme}

Now that we have developed a couple of basic behavior blocks,
we can use them to compose the overall behavior for automated driving,
as shown in~\cref{fig:03_2_arbitration_graph}, starting bottom-up.

We follow a similar notation to~\cite{lauer_cognitive_2010},
denoting the behavior options of an arbitrator with $O_\textsc{ArbitratorName}$,
using round brackets \enquote{$()$} for an ordered list and curly brackets \enquote{$\{\}$} for a set of options.
Basic behavior blocks are highlighted with \textsl{ItalicNames} and arbitrators with \textsc{CapitalNames}.

In an urban environment possible behaviors are \textsl{Follow\-EgoLane},
\textsl{ChangeLane}, \textsc{MergeIntoLane} and \textsc{CrossIntersection}.
In order to clear intersections as soon as reasonably possible and not to change lanes unintentionally in an intersection,
\textsc{CrossIntersection} has clear priority at intersections.
The remaining urban behaviors typically have no clear and consistent priority over each other though ---
yet the most reasonable one should be chosen.

As none of the existing arbitration schemes (by priority, sequence or random) are sufficient for this task,
we define a new cost-based arbitrator that selects the behavior option with the lowest expected cost.
A hysteresis prevents oscillating behavior choices.
By introducing cost arbitrators, the decision-making concept can be extended to dynamically changing preferences.

However, cost arbitrators should be used with care.
First of all, the cost estimates of an arbitrators behavior options have to be comparable.
This could easily lead to cross-dependencies of behavior blocks.
Secondly, if the cost contains too many obfuscated objectives, the selection process becomes difficult to understand.
Both are properties we actually want to avoid.
Therefore, we advise to use cost arbitrators rarely and with simple, generic costs.
In our case, we use a simple estimate of the expected travel velocity:
\begin{align*}
  O_\textsc{UrbanDriving} =   \text{\faDollar}\hspace{.1em}
                              \{&\textsl{FollowEgoLane},\\
                              &\textsl{ChangeLaneLeft / -Right},\\
                              &\textsc{MergeIntoLaneLeft / -Right}\}
\end{align*}

As discussed in~\cref{subsec:03_main_2_automated_driving_3_driving_maneuvers}
lane changes in dense traffic can be decomposed into three stages.
As a result, a sequence-based arbitrator is used to compose \textsc{MergeIntoLane}:
\begin{align*}
  O_\textsc{MergeIntoLane} = \text{\faTasks}\hspace{.1em}
                             (  &\textsl{ApproachGap}, \textsl{IndicateIntention},\\
                                &\textsl{MergeIntoGap})
\end{align*}

Highway behaviors are combined using a cost arbitrator:
\begin{align*}
  O_\textsc{HighwayDriving} = \text{\faDollar}\hspace{.1em}
                              \{  &\textsl{MergeOntoHighway},\\
                                  &\textsl{FollowHighwayLane},\\
                                  &\textsl{ChangeHighwayLaneLeft / -Right},\\
                                  &\textsl{ExitFromHighway}\}
\end{align*}

In case of \textsc{Parking} at most one option is feasible after all,
such that a trivial priority-based arbitrator can be used:
\begin{align*}
  O_\textsc{Parking} = \text{\faSortNumericAsc}\hspace{.1em} (\textsl{LeaveGarage}, \textsl{ParkNearGoal})
\end{align*}

The emergency maneuvers for unavoidable collisions are grouped together using a
cost-based arbitrator estimating the expected damage.
In such a way, it chooses the option with the lowest expected damage:
\begin{align*}
  O_\textsc{AvoidCollisionInLastResort} = \text{\faDollar}\hspace{.1em} \{\textsl{EmergenyStop}, \textsl{EvadeObject}\}
\end{align*}

\begin{figure*}
  \hspace{-4em}%
\begingroup
  \makeatletter
  \providecommand\color[2][]{%
    \GenericError{(gnuplot) \space\space\space\@spaces}{%
      Package color not loaded in conjunction with
      terminal option `colourtext'%
    }{See the gnuplot documentation for explanation.%
    }{Either use 'blacktext' in gnuplot or load the package
      color.sty in LaTeX.}%
    \renewcommand\color[2][]{}%
  }%
  \providecommand\includegraphics[2][]{%
    \GenericError{(gnuplot) \space\space\space\@spaces}{%
      Package graphicx or graphics not loaded%
    }{See the gnuplot documentation for explanation.%
    }{The gnuplot epslatex terminal needs graphicx.sty or graphics.sty.}%
    \renewcommand\includegraphics[2][]{}%
  }%
  \providecommand\rotatebox[2]{#2}%
  \@ifundefined{ifGPcolor}{%
    \newif\ifGPcolor
    \GPcolortrue
  }{}%
  \@ifundefined{ifGPblacktext}{%
    \newif\ifGPblacktext
    \GPblacktextfalse
  }{}%
  \let\gplgaddtomacro\g@addto@macro
  \gdef\gplbacktext{}%
  \gdef\gplfronttext{}%
  \makeatother
  \ifGPblacktext
    \def\colorrgb#1{}%
    \def\colorgray#1{}%
  \else
    \ifGPcolor
      \def\colorrgb#1{\color[rgb]{#1}}%
      \def\colorgray#1{\color[gray]{#1}}%
      \expandafter\def\csname LTw\endcsname{\color{white}}%
      \expandafter\def\csname LTb\endcsname{\color{black}}%
      \expandafter\def\csname LTa\endcsname{\color{black}}%
      \expandafter\def\csname LT0\endcsname{\color[rgb]{1,0,0}}%
      \expandafter\def\csname LT1\endcsname{\color[rgb]{0,1,0}}%
      \expandafter\def\csname LT2\endcsname{\color[rgb]{0,0,1}}%
      \expandafter\def\csname LT3\endcsname{\color[rgb]{1,0,1}}%
      \expandafter\def\csname LT4\endcsname{\color[rgb]{0,1,1}}%
      \expandafter\def\csname LT5\endcsname{\color[rgb]{1,1,0}}%
      \expandafter\def\csname LT6\endcsname{\color[rgb]{0,0,0}}%
      \expandafter\def\csname LT7\endcsname{\color[rgb]{1,0.3,0}}%
      \expandafter\def\csname LT8\endcsname{\color[rgb]{0.5,0.5,0.5}}%
    \else
      \def\colorrgb#1{\color{black}}%
      \def\colorgray#1{\color[gray]{#1}}%
      \expandafter\def\csname LTw\endcsname{\color{white}}%
      \expandafter\def\csname LTb\endcsname{\color{black}}%
      \expandafter\def\csname LTa\endcsname{\color{black}}%
      \expandafter\def\csname LT0\endcsname{\color{black}}%
      \expandafter\def\csname LT1\endcsname{\color{black}}%
      \expandafter\def\csname LT2\endcsname{\color{black}}%
      \expandafter\def\csname LT3\endcsname{\color{black}}%
      \expandafter\def\csname LT4\endcsname{\color{black}}%
      \expandafter\def\csname LT5\endcsname{\color{black}}%
      \expandafter\def\csname LT6\endcsname{\color{black}}%
      \expandafter\def\csname LT7\endcsname{\color{black}}%
      \expandafter\def\csname LT8\endcsname{\color{black}}%
    \fi
  \fi
    \setlength{\unitlength}{0.0500bp}%
    \ifx\gptboxheight\undefined%
      \newlength{\gptboxheight}%
      \newlength{\gptboxwidth}%
      \newsavebox{\gptboxtext}%
    \fi%
    \setlength{\fboxrule}{0.5pt}%
    \setlength{\fboxsep}{1pt}%
\begin{picture}(10770.00,2266.00)%
    \gplgaddtomacro\gplbacktext{%
      \colorrgb{0.50,0.50,0.50}%
      \put(2178,751){\makebox(0,0)[r]{\strut{}SafeStop}}%
      \colorrgb{0.50,0.50,0.50}%
      \put(2178,1075){\makebox(0,0)[r]{\strut{}ChangeLaneRight}}%
      \colorrgb{0.50,0.50,0.50}%
      \put(2178,1398){\makebox(0,0)[r]{\strut{}FollowEgoLane}}%
      \colorrgb{0.50,0.50,0.50}%
      \put(2178,1722){\makebox(0,0)[r]{\strut{}ChangeLaneLeft}}%
      \colorrgb{0.50,0.50,0.50}%
      \put(2178,2045){\makebox(0,0)[r]{\strut{}ParkNearGoal}}%
      \colorrgb{0.50,0.50,0.50}%
      \put(2357,484){\makebox(0,0){\strut{}0}}%
      \colorrgb{0.50,0.50,0.50}%
      \put(3693,484){\makebox(0,0){\strut{}100}}%
      \colorrgb{0.50,0.50,0.50}%
      \put(5029,484){\makebox(0,0){\strut{}200}}%
      \colorrgb{0.50,0.50,0.50}%
      \put(6365,484){\makebox(0,0){\strut{}300}}%
      \colorrgb{0.50,0.50,0.50}%
      \put(7701,484){\makebox(0,0){\strut{}400}}%
      \colorrgb{0.50,0.50,0.50}%
      \put(9037,484){\makebox(0,0){\strut{}500}}%
      \colorrgb{0.50,0.50,0.50}%
      \put(10373,484){\makebox(0,0){\strut{}600}}%
    }%
    \gplgaddtomacro\gplfronttext{%
      \csname LTb\endcsname%
      \put(6365,154){\makebox(0,0){\strut{}Time [s]}}%
    }%
    \gplbacktext
    \put(0,0){\includegraphics{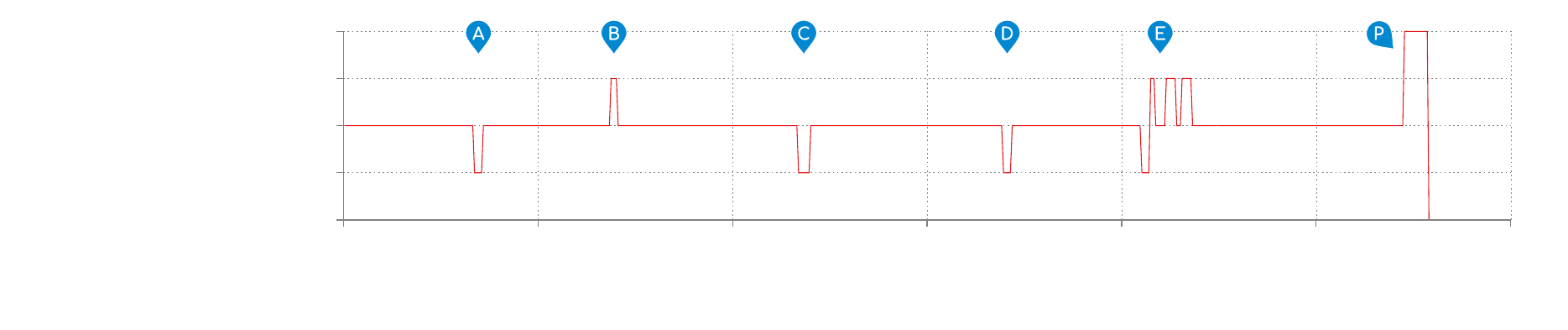}}%
    \gplfronttext
  \end{picture}%
\endgroup
   \caption{
    Behavior choices in the experiment driving the whole test track.
  }\label{fig:04_results}
\end{figure*}
\begin{figure}
  \centering
  \includegraphics[width=\columnwidth]{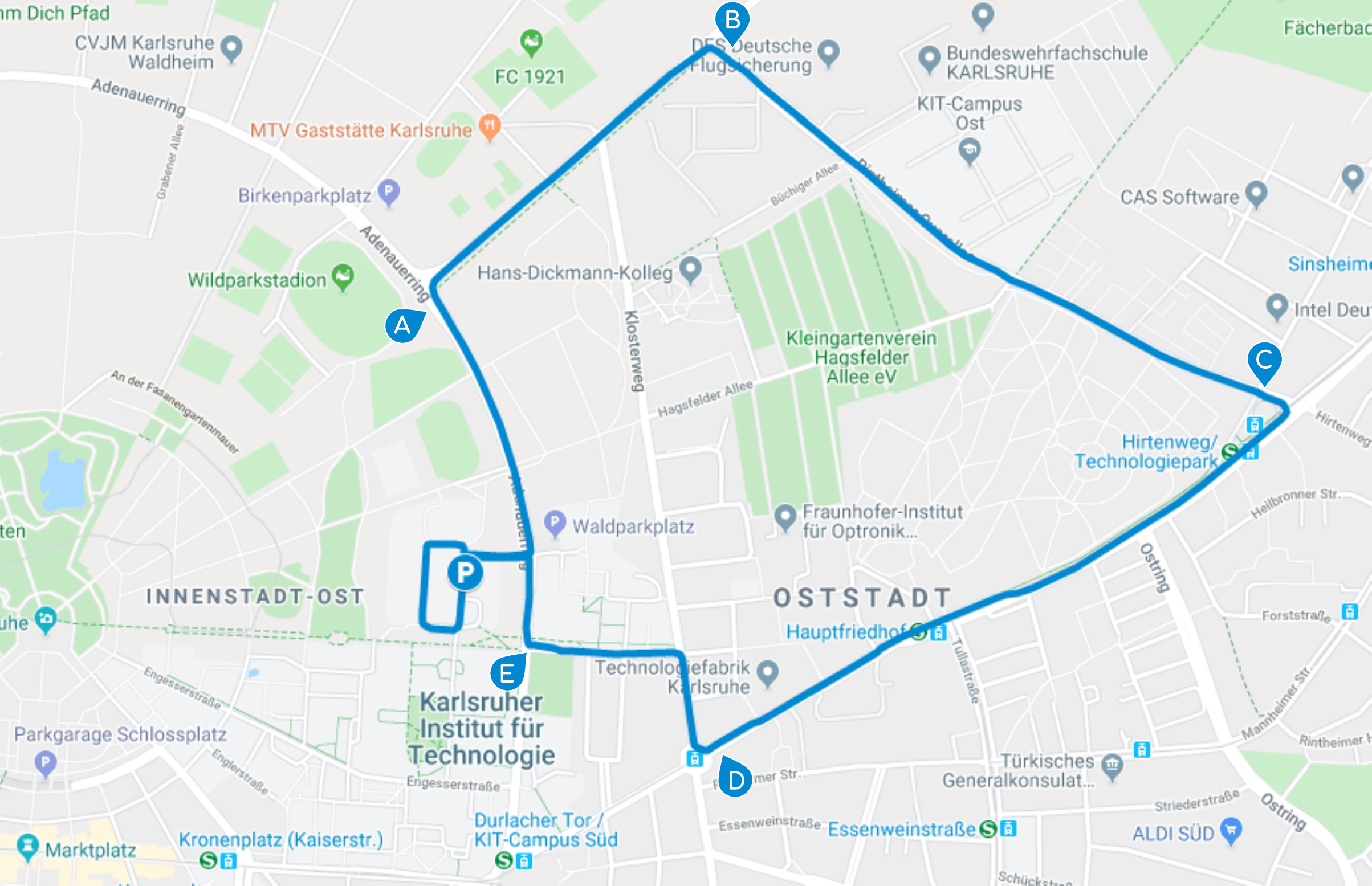}
  \caption{
    Test track running $\SI{5.7}{\kilo\meter}$ through Karlsruhe, Germany.
    Start and end position is a parking lot on the university campus.
    Tiles \copyright~2020 Google, Map data \copyright~2020 GeoBasis-DE/BKG.%
  }\label{fig:04_track}
\end{figure}
\begin{figure}
  \centering
  \includegraphics[width=.9\columnwidth]{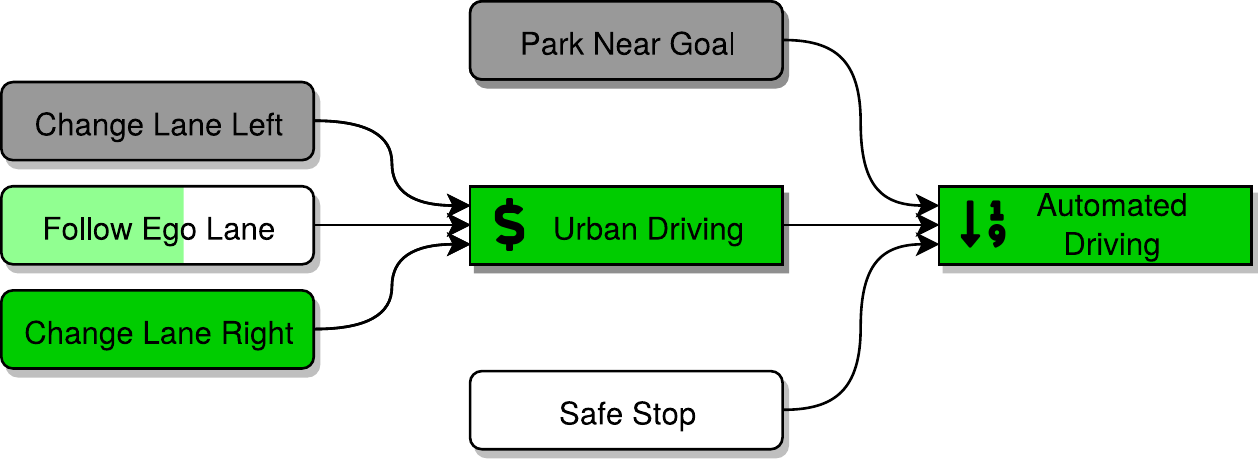}
  \caption{
    Example arbitration graph, as used in our simulative experiments.
    Colors depict the state at point~$E$:
    Grey: invocation condition false,
    dark green: active behavior branch,
    light green: utility (normalized inverse costs, see also \cref{fig:04_follow_vs_change_lane}).
  }\label{fig:04_arbitration_graph}
\end{figure}
 
Finally, these arbitrators and the \textsl{SafeStop} fallback
are composed together to the top-most priority-based arbitrator:
\begin{align*}
  O_\textsc{AutomatedDriving} = \text{\faSortNumericAsc}\hspace{.1em}
                                ( &\textsc{AvoidCollisionInLastResort},\\
                                  &\textsc{Parking}, \textsc{CrossIntersection},\\
                                  &\textsc{UrbanDriving},\\
                                  &\textsc{HighwayDriving}, \textsl{SafeStop})
\end{align*}

\section{Experiments}%
\label{sec:04_experiments}

In this section, we show the applicability of the proposed concept to utilize a
hierarchical behavior-based architecture for behavior generation in automated driving.

\subsection{Setup}%
\label{subsec:04_experiments_1_setup}

The explanatory example performs basic urban driving behaviors on a simulated $\SI{5.7}{\kilo\meter}$ test track
based on our real-world test route in Karlsruhe, Germany.
The route, shown in~\cref{fig:04_track}, contains segments with speed limits of $\SI{30}{\kmh}$, $\SI{50}{\kmh}$ and $\SI{60}{\kmh}$,
is crossing or turning at 12 intersections, traversing one roundabout and ends at a parking lot.

We use the ROS-based open-source simulation framework CoInCar-Sim~\cite{naumann_coincar-sim_2018}.
One great advantage of this framework is that it provides the same interface as our test vehicle Bertha~\cite{ziegler_making_2014}.
Hence, we can develop, test and deploy the same behavior and planning pipeline in CoInCar-Sim and Bertha.

Our basic example maneuvers for this track are:
\textsl{ParkNearGoal}, \textsl{FollowEgoLane}, \textsl{ChangeLane} (one instantiation for left, another for right lane changes) and \textsl{SafeStop}.
Lane following and both lane change behaviors are combined within a cost-based \textsc{UrbanDriving} arbitrator.
Whereas parking, urban driving and the safe stop fallback constitute the overall behavior
using a priority-based \textsc{AutomatedDriving} arbitrator.
\Cref{fig:04_arbitration_graph} illustrates this arbitration graph.

This design has the following motivation.
\textsl{ParkNearGoal} is only applicable in the vicinity of the goal and a nearby parking lot.
Thus, as long as the ego vehicle is still on the route \textsl{FollowEgoLane} is
and \textsl{ChangeLaneLeft} or \textsl{ChangeLaneRight} might be applicable.
\textsc{UrbanDriving} will select the most promising one, w.r.t.\ the expected average velocity,
routing costs and lane change penalties.
As soon as the vehicle approaches its goal, \textsl{FollowEgoLane} will bring it to a stop within the last lanelet.
Then, \textsl{ParkNearGoal} will become applicable, chosen by priority and lead the car into its parking lot.
When the parking maneuver is finished, \textsl{ParkNearGoal} will render inapplicable again.
At that point also none of the \textsc{UrbanDriving} behaviors are applicable any more because the car has left the route.
As a result \textsc{AutomatedDriving} selects the lowest priority behavior \textsl{SafeStop}.
This is a good illustration of how the fallback behavior
prevents undefined states and keeps the vehicle in a safe position.

\subsection{Results}%
\label{subsec:04_experiments_2_results}

\Cref{fig:04_results} shows the resulting behavior selection over time.
The whole route takes 9:40min and features the expected behavior characteristics.
The vehicle starts leaving the campus area by following the lane.
At intersection $A$, it changes to the right lane in order to take a turn into a north-east direction.
At point $B$, it takes another right turn following the ego lane and has to change to the left lane.
When approaching the next intersection $C$, the ego vehicle changes onto the exit lane in order to turn into south-east direction.
At $t=\SI{339}{\second}$ it approaches and passes the roundabout $D$.

\begin{figure}
  \centering
  \small\footnotesize
  \def\svgwidth{.8\columnwidth}%
\begingroup%
  \makeatletter%
  \providecommand\color[2][]{%
    \errmessage{(Inkscape) Color is used for the text in Inkscape, but the package 'color.sty' is not loaded}%
    \renewcommand\color[2][]{}%
  }%
  \providecommand\transparent[1]{%
    \errmessage{(Inkscape) Transparency is used (non-zero) for the text in Inkscape, but the package 'transparent.sty' is not loaded}%
    \renewcommand\transparent[1]{}%
  }%
  \providecommand\rotatebox[2]{#2}%
  \newcommand*\fsize{\dimexpr\f@size pt\relax}%
  \newcommand*\lineheight[1]{\fontsize{\fsize}{#1\fsize}\selectfont}%
  \ifx\svgwidth\undefined%
    \setlength{\unitlength}{1030.00000961bp}%
    \ifx\svgscale\undefined%
      \relax%
    \else%
      \setlength{\unitlength}{\unitlength * \real{\svgscale}}%
    \fi%
  \else%
    \setlength{\unitlength}{\svgwidth}%
  \fi%
  \global\let\svgwidth\undefined%
  \global\let\svgscale\undefined%
  \makeatother%
  \begin{picture}(1,0.815534)%
    \lineheight{1}%
    \setlength\tabcolsep{0pt}%
    \put(0,0){\includegraphics[width=\unitlength,page=1]{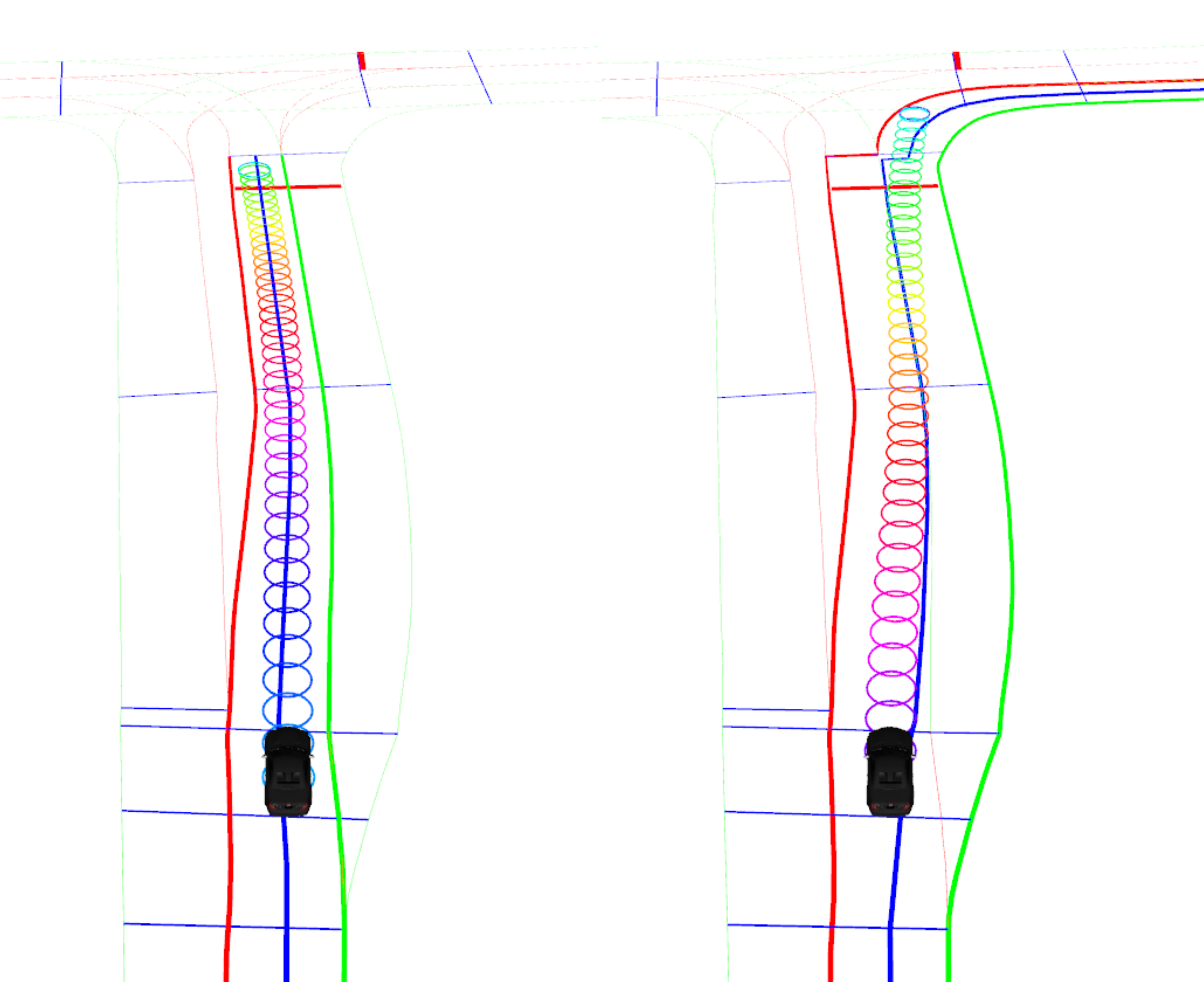}}%
    \put(0.01317755,0.59720356){\color[rgb]{0,0,0}\makebox(0,0)[t]{\lineheight{1.25}\smash{\begin{tabular}[t]{c}\textit{FollowEgoLane:}\end{tabular}}}}%
    \put(0.50188704,0.59720389){\color[rgb]{0,0,0}\makebox(0,0)[t]{\lineheight{1.25}\smash{\begin{tabular}[t]{c}\textit{ChangeLaneRight:}\end{tabular}}}}%
  \end{picture}%
\endgroup%
   \caption{
    \textsl{FollowEgoLane} and \textsl{ChangeLaneRight} maneuver corridors at point $E$.
    The route continues to the right at this point.
    As a result, the \textsl{FollowEgoLane} corridor ends in $\SI{74}{\meter}$, while the \textsl{ChangeLaneRight} corridor has a length of $\SI{243}{\meter}$.
  }\label{fig:04_follow_vs_change_lane}
\end{figure}

\Cref{fig:04_follow_vs_change_lane} shows the two applicable behavior options at point $E$, where the route leads onto the \enquote{Adenauerring} again.
The route continues with a right turn from the rightmost lane, while the ego is on the leftmost lane still.
This is a suitable scenario to explain the cost-based arbitration in detail.
The urban driving cost estimate incorporates the average expected travel velocity, routing costs and penalizes lane changes:
\begin{align*}
  J &= -\hat{v} + n_\text{LCNeeded} \cdot J_\text{LCNeeded},   \quad\quad\text{without lane change}\\
  J &= -\hat{v} + n_\text{LCNeeded} \cdot J_\text{LCNeeded} + J_\text{LCManeuver},   \quad\text{otherwise}
\end{align*}
As a simple, yet effective heuristic, we estimate $\hat{v}$, the expected average velocity of this maneuver,
from the maneuver corridor length and speed limit as shown in \cref{fig:04_follow_vs_change_lane}.
For routing, we charge each lane change needed to follow the route after this command with $J_\text{LCNeeded}=\SI{10}{\kmh}$.
Lane change behaviors themselves are penalized with a lower $J_\text{LCManeuver}=\SI{5}{\kmh}$.
Hence, the arbitrator generally prefers the follow lane behavior as long as it matches the route.
As soon as one or multiple lane changes will be necessary, this maneuver will become more favorable.

At point $E$, the behaviors have these costs:%
\begin{align*}
  J_\textsl{FollowEgoLane}    &= -25.0 + 1 \cdot 10.0       &= -15.0 \\
  J_\textsl{ChangeLaneRight}  &= -33.4 + 0 \cdot 10.0 + 5.0 &= -28.4
\end{align*}
Consequently the cost-based arbitrator chooses \textsl{ChangeLaneRight}, which has lower cost than \textsl{FollowEgoLane},
as also illustrated in \cref{fig:04_arbitration_graph}.

An interesting part is directly after taking the right turn at point $E$ from $t=\SI{422}{\second}$ to $t=\SI{436}{\second}$.
Here, the vehicle performs two consecutive lane changes in order to pass this two-lane road from the rightmost lane to the exit lane.
This is especially noteworthy, as no double lane change or other hand-crafted behavior has been defined for such a scenario.
The behavior emerges purely because the routing has been incorporated into the cost estimate.

The road leads back to the campus again, where the vehicle slows down and stops at the end of the route.
Finally, the parking behavior becomes active and brings the car into its parking lot.
After finishing the parking maneuver, the safe stop behavior is the last suitable option and keeps the car at a standstill.

Please also consider our video: \href{https://youtu.be/qdIwchDGA_g}{youtu.be/qdIwchDGA\_g}

\section{Conclusions and Future Work}%
\label{sec:05_conclusion}

This publication presented the following contributions:

An extension to the hierarchical behavior-based arbitration concept proposed in~\cite{lauer_cognitive_2010}.
We introduced a cost-based arbitration scheme that is helpful when multiple behavior options are applicable
but have no clear and consistent priority among each other.

We have formulated a behavior generation stack for AVs based on the hierarchical behavior-based arbitration scheme.
It consists of maneuvers for urban and highway environments, contains parking and emergency behaviors,
and prevents undefined states with a fallback safe stop behavior.

We have shown the usefulness and applicability of our design in an explanatory evaluation on a simulated route.

The key advantages of the approach are:

\begin{itemize}
  \item \textbf{Scenario-specific solutions can be combined easily.}\\
    In the experiments, five different behaviors have been employed to handle various scenarios,
    from four-way intersections, T-junctions, a roundabout to multi-lane bypass roads and parking.

  \item \textbf{It supports different planning approaches.}\\
    We utilized two different trajectory planners in our experiments.
    Urban corridor-based maneuvers used an optimization-based planner similar to~\cite{ziegler_trajectory_2014},
    while the parking maneuver generated Hybrid Curvature trajectories with an RRT* motion planner~\cite{banzhaf_footprints_2018}.
    But also different approaches could be used for the same behavior.

  \item \textbf{The resulting behavior can be well explained.}\\
    The strongly modular design significantly improves understandability compared to FSMs or classical behavior-based systems.
    Each invocation condition can be well understood; the selection logic of arbitrators is comprehensive.
    As a result, the hierarchical decision-making process can be well explained and traced over time.

  \item \textbf{It can be iteratively extended by more behaviors.}\\
    In order to add the parking behavior to our behavior generation,
    the definition of its invocation and commitment conditions was sufficient to add it to the \textsc{AutomatedDriving} arbitrator.
    Thanks to the strong decoupling, no changes to any other behavior block were necessary.

  \item \textbf{The modularity supports robustness and efficiency.}\\
    Each of the behavior blocks is self-contained, such that occurring failures are contained as well and do not affect the overall system stability.
    In case of a failure, the system will degrade seamlessly by ignoring this behavior option.
    Furthermore, the atomic structure allows to evaluate behavior options in parallel to increase efficiency.
    Strong modularity has many more advantages, among others, reusability and maintainability.

  \item \textbf{Complex behavior emerges from simple components.}\\
    Complex system behavior, as multiple consecutive lane changes to approach an exit lane, %
    emerges from the arbitration scheme without the need for hand-crafted decision or planning logic.
\end{itemize}

These benefits have led to a smooth development process with promising results, as outlined in~\cref{sec:04_experiments}.
Thus, we look forward to further enhance the numerous existing behavior blocks,
extend the behavior stack by e.g.\ our MIQP approach for cooperative zip merges~\cite{burger_cooperative_2018}
and most excitingly to integrate this stack on our test vehicle Bertha.

\printbibliography

\end{document}